\documentclass{article}

\usepackage{arxiv}
\usepackage{graphicx}
\usepackage[utf8]{inputenc} % allow utf-8 input
\usepackage[T1]{fontenc}    % use 8-bit T1 fonts
\usepackage{hyperref}       % hyperlinks
\usepackage{url}            % simple URL typesetting
\usepackage{booktabs}       % professional-quality tables
\usepackage{amsfonts}       % blackboard math symbols
\usepackage{nicefrac}       % compact symbols for 1/2, etc.
\usepackage{microtype}      % microtypography
\usepackage{lipsum}

\title{Utilizing Differential Evolution into optimizing targeted cancer treatments}

\author{
  Michail-Antisthenis Tsompanas, Larry Bull, Andrew Adamatzky \\
  Unconventional Computing Laboratory \\
   Department of Computer Science and Creative Technologies \\
    University of the West of England\\
   Bristol, UK
     \And
     Igor Balaz\\
     Laboratory for Meteorology, Physics and Biophysics \\ 
 Faculty of Agriculture \\
University of Novi Sad\\
Novi Sad, Serbia
}

\begin{document}
\maketitle

\begin{abstract}
Working towards the development of an evolvable cancer treatment simulator, the investigation of Differential Evolution was considered, motivated by the high efficiency of variations of this technique in real-valued problems. A basic DE algorithm, namely ``DE/rand/1" was used to optimize the simulated design of a targeted drug delivery system for tumor treatment on PhysiCell simulator. The suggested approach proved to be more efficient than a standard genetic algorithm, which was not able to escape local minima after a predefined number of generations. The key attribute of DE that enables it to outperform standard EAs, is the fact that it keeps the diversity of the population high, throughout all the generations. This work will be incorporated with ongoing research in a more wide applicability platform that will design, develop and evaluate targeted drug delivery systems aiming cancer tumours.
\end{abstract}

% keywords can be removed

%%%%%%%%%%%%%%%%%%%%%%%%%%%%%%%%%%%%%%%%%%%%%%%%%%%%%%%%%%%%%%%%%

\section{Introduction}
\label{S:1}

Differential evolution (DE) gained popularity over other well-established evolutionary algorithms (EAs), as it follows similar algorithmic steps with standard EAs, but was able to surpass them in terms of efficiency \cite{storn1997differential}. There have been several proposals on how to enhance its performance \cite{thomsen2004multimodal,epitropakis2011enhancing} and these alternative algorithmic approaches, building on the initial methodology, were tested in real problems , as well as numerical benchmark problems \cite{qin2008differential,das2010differential,DAS20161,wang2013gaussian}.

DE was initially proposed in \cite{storn1997differential}, as an effective methodology of optimization over continuous spaces of nonlinear and non differentiable functions. The method was introduced as an alternative to previous direct search approaches, aiming at three main objectives: the ability to find the global optimum, the fast convergence to this optimum and the need of a small amount of control parameters for the procedure. Moreover, it was developed to be easily implemented in parallel computing platforms. The methodology is population based, where for every generation the new individuals are produced after applying the scaled difference (hence the name differential evolution) of some predefined individuals to another predefined base individual. %If the fitness of the new individual () is better than one predecessor, the new individual replaces its predecessor in the evolved population. 

The ability to easily extract attributes of the population, such as the distance of its members and their directions, through the aforementioned methodology, is what makes DE so powerful. This characteristic is defined as self-referential mutation \cite{price1997differential}. Given the aforementioned advantages of DE compared with other EAs, it was chosen to investigate the optimization of a targeted drug delivery system (DDS) on a cancer tumour, when simulated with PhysiCell \cite{ghaffarizadeh2018physicell}.

PhysiCell \cite{ghaffarizadeh2018physicell} is a multicellular, agent-based simulator that was designed to extend the BioFVM \cite{ghaffarizadeh2015biofvm} framework, to form a virtual laboratory. PhysiCell is open source and offers several sample projects, one of which is the one studied in this study. More specifically, sample project ``anti-cancer biorobots" \cite{ghaffarizadeh2018physicell} was developed as a possible tool to investigate the targeted cancer treatment, i.e. with drugs transported by specialized nanoparticles that would target specific cells of the cancer tumours. 

Previously, PhysiCell was deployed as a virtual laboratory in the optimization process of the design of nanoparticle carriers of cancer treating compounds \cite{PREEN20191,tsompanas2019} and the process of mapping immunotherapies \cite{ozik2019learning}. More specifically, the use of PhysiCell led the training of surrogate-assisted evolutionary algorithms exploring the efficient solutions of the design of nanoparticle-based drug delivery systems for cancer \cite{PREEN20191}. A similar application of PhysiCell was examined in \cite{tsompanas2019}, where the optimization was achieved by a novel memetic algorithm inspired by the fundamental haploid-diploid lifecycle of eukaryotic organisms \cite{bull1999baldwin,bull2017evolution}. A similar application is considered here, however the exploration of the solution space is investigated through DE. Moreover, active learning and genetic algorithms incorporated with the PhysiCell simulator, enabled the efficient exploration of biological and clinical constraints for cancer immunotherapy \cite{ozik2019learning}.

\section{Differential evolution}
\label{S:2}

The methodology of DE comprises of the following steps: initialization, mutation, recombination (or crossover) and selection. At first, a population of individuals (possible solutions) is formed by randomly picking values on the D-dimensional search space of the problem to be optimized. The random function used is mainly uniformly distributed to cover sufficiently the search space, which could be limited with lower and upper boundaries depending on the definition of the problem.

Then, the mutation operator is employed on the initial population. More specifically, for every individual in the population a new individual is generated (named the mutant individual) by a mathematical expression of the parameters of randomly chosen or predefined individuals. For example, the mutant individual ($\mathsf{v}_i$) can be the linear combination of three randomly selected individuals as defined by the following equation, where $x_{r1}$, $x_{r2}$ and $x_{r3}$ are randomly selected individuals from the population and $F$ is a scaling factor, which is a positive number.

\begin{equation}
\label{eq:mut1}
\mathsf{v}_i = x_{r1}+F\cdot (x_{r2}-x_{r3})
\end{equation}

Note here, that the variations of DE methodology are defined with a notation in the form of ``DE/base/num". The ``base" part of the notation refers to the technique of choosing the individual that will be used as the base individual to which the scaled difference will be added. The ``num" part indicates the amount of pairs of individuals that will produce a scaled difference each, to be added on the base individual. %The ``cross" part reveals the manner that the crossover operator is implemented. \textbf{Here, the ``bin" notation is used as all the implemented algorithms utilize the binomial method of crossover.} 
Thus, the aforementioned DE variation is denoted as ``DE/rand/1".

The produced mutant individuals, then, undergo the crossover operator, in order to be recombined with individuals from the initial population. Each individual chosen from the initial population is denoted as the target individual and the produced individual after the crossover as the trial individual. Two crossover strategies are mainly used, namely exponential and binomial. However, binomial is dominant in the literature, where every parameter in the D-dimensional solution is treated separately to the others, and chosen from the mutant or the target individual based on a probability defined as the $CR$ parameter. This parameter is known as the crossover rate.

The final step of the algorithm is then evaluation, by the use of the selection operator. Similar to standard GAs, the trial individuals produced from the previous step, are evaluated with the fitness function. If their fitness is better than the one of their target individual, they replace the target individual in the next generation. Otherwise the target individual is retained. When all trial individuals are tested and the appropriate selection is made to form the population of the next generation, the algorithm runs again the population through the mutation, crossover and selection operators, until the termination criteria are met (i.e. the computation budget).

\section{Methodology}
\label{S:3}

\begin{figure}[!t]
\centering
\includegraphics[width=0.5\linewidth]{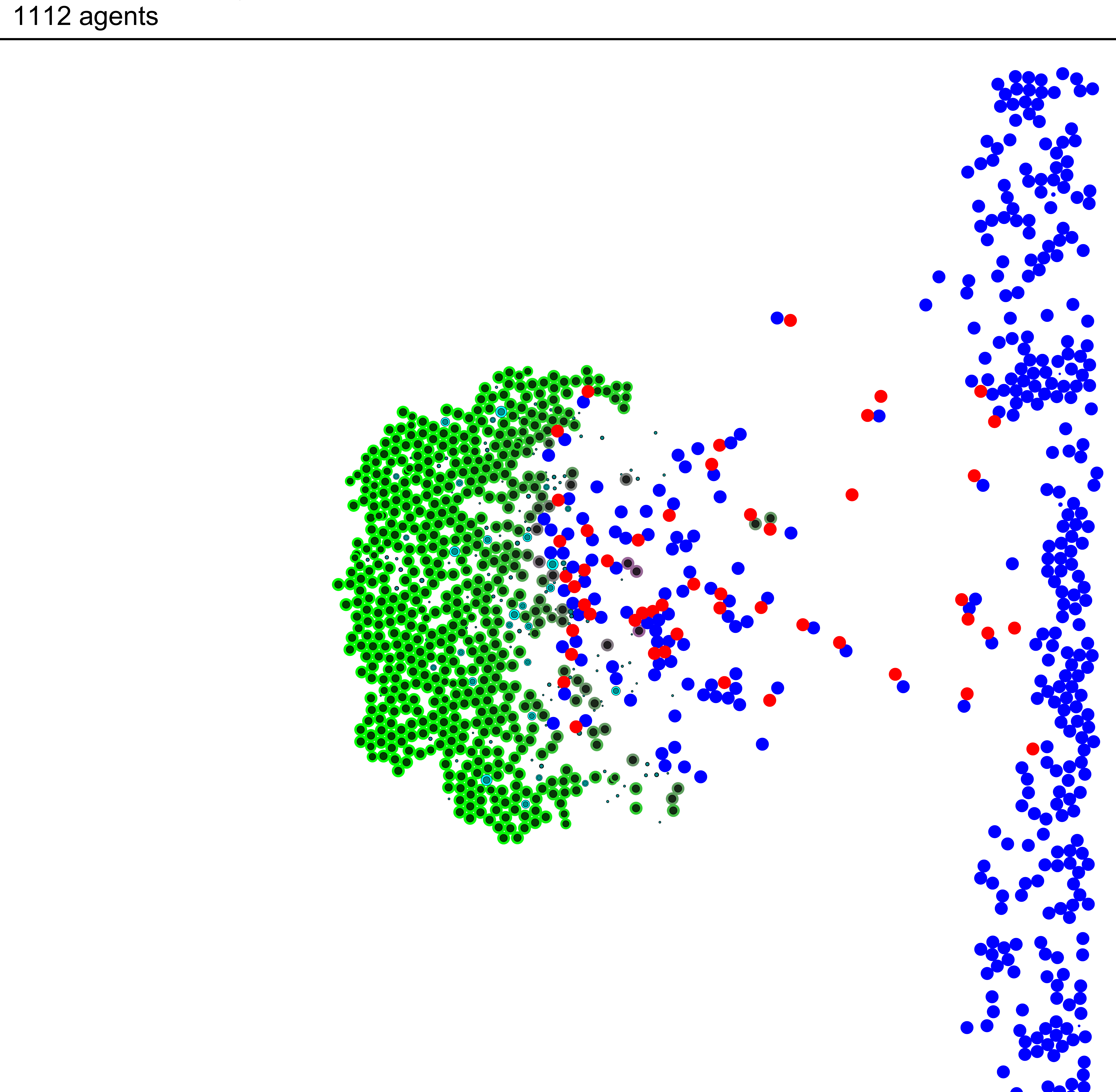}
\caption{The graphical representation output of PhysiCell after 10 days of simulated growth and treatment of a cancer tumour.}
\label{tumour}
\end{figure}

The sample project from PhysiCell framework that was investigated, is the “anti-cancer biorobots". In this simulation all entities are simulated as agents. Namely, there are three different types of agents with different functionalities, the cancer cells, the chemical compound (defined as cargo agents) and the functionalized nanoparticles (defined as worker agents). The outputs of PhysiCell include graphical representation of the agents in the simulated areas. An example after 10 days of simulated growth and treatment of a tumour is depicted in Fig. \ref{tumour}. The cancer cells are illustrated as green, the chemical compound as blue, while the nanoparticles as red agents.

The problem to be optimized here was defined as the design of nanoparticle agents in PhysiCell simulator (v.1.4.1) that will result in lower number of remaining cancer cells. Each possible solution is mapped in a 6-dimensional space where the parameters studied and their boundaries are the attached worker migration bias [0,1], the unattached worker migration bias [0,1], worker relative adhesion [0,10], worker relative repulsion [0,10], worker motility persistence time (min) [0,10] and the cargo release $O_2$ threshold (mmHg) [0,20]. Due to the stochastic nature of the simulator each possible solution is evaluated after extracting the average value of the remaining cancer cells of 5 runs. Each run executes 7 days of growing an initial 200 micron radius tumor and 3 days of applying the treatment.

For comparison reasons, a generic GA was applied to optimize the aforementioned problem. The parameters of the GA were chosen as population size $P=20$, tournament size $T=2$ for selection and replacement operations, uniform crossover with probability $X=80\%$ and per allele mutation rate of $\mu=20\%$ with random step size of $s=[-5,5]\%$.

On the other hand, the ``DE/rand/1" strategy was implemented and tested for the optimization of a targeted DDS on a cancer tumour, by simulating this procedure with PhysiCell. % Note that all six aforementioned variations were implemented, as well as the variations of Crowding-based DE \cite{thomsen2004multimodal} and Proximity-based DE \cite{epitropakis2011enhancing}. Analyzing the results from these algorithms (all variations result in 18 algorithms), remains as an aspect of future work.
The parameters of the DE algorithm were chosen as population size $P=20$, scaling factor $F=0.5$ and crossover rate $CR=0.9$. Note that these parameters are not fine-tuned to enhance the performance of the algorithm, but are most commonly used throughout the literature.

The rest of the parameters used to define the simulation of the tumour environment by PhysiCell were retained unchanged. More specifically these parameters are listed in Table \ref{tabl:1}.

\begin{table}
\centering
\caption{Unaltered parameters of PhysiCell simulator.}
\label{tabl:1}       % Give a unique label
\begin{tabular}{|l|l|}
\noalign{\smallskip}\hline
Parameter & Value  \\\hline
\noalign{\smallskip}\noalign{\smallskip}\hline
Damage rate &  0.03333 $min^{-1}$\\ \hline
Repair rate &  0.004167 $min^{-1}$\\ \hline
Drug death rate &  0.004167 $min^{-1}$\\ \hline
Elastic coefficient &  0.05 $min^{-1}$  \\ \hline
Cargo $O_2$ relative uptake   & 0.1 $min^{-1}$ \\ \hline
Cargo apoptosis rate       & 4.065e-5 $min^{-1}$ \\  \hline
Cargo relative adhesion    & 0  \\ \hline
Cargo relative repulsion   & 5  \\ \hline
Maximum relative cell adhesion distance & 1.25 \\  \hline
Maximum elastic displacement & 50 $\mu m$  \\ \hline
Maximum attachment distance & 18 $\mu m$ \\ \hline
Minimum attachment distance & 14 $\mu m$ \\  \hline
Motility shutdown detection threshold &  0.001 \\ \hline
Attachment receptor threshold  & 0.1 \\ \hline
Worker migration speed     & 2 $\mu m / min$ \\  \hline
Worker apoptosis rate      & 0 $min^{-1}$ \\  \hline
Worker $O_2$ relative uptake  & 0.1 $min^{-1}$ \\ \hline
\noalign{\smallskip}
\end{tabular}
\end{table}

\section{Results}
\label{S:4}

\begin{figure}[!t]
\centering
\includegraphics[width=0.5\linewidth]{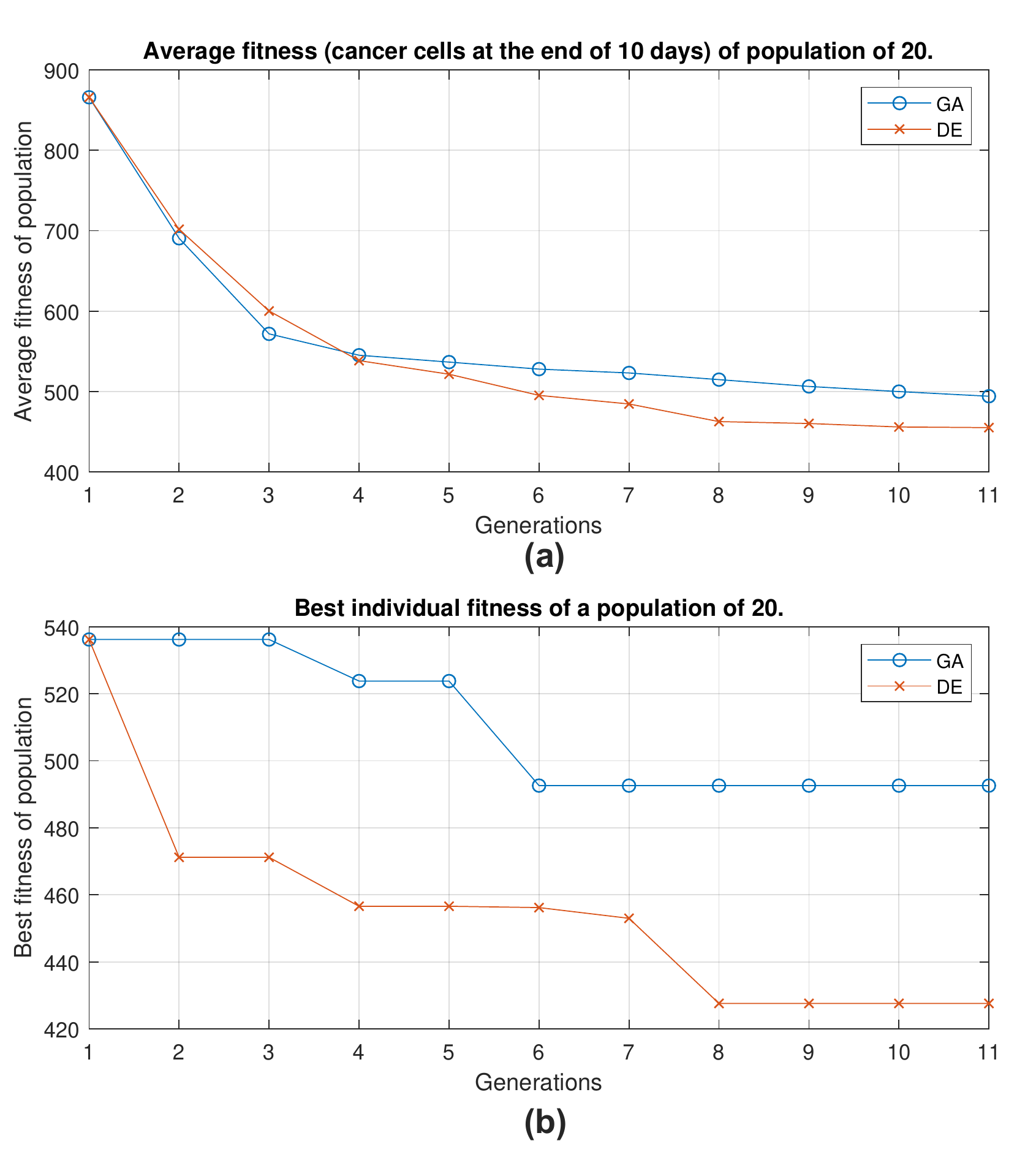}
\caption{Average (a) and best fitness (b) of the populations evolved with GA and DE in the first comparison run.}
\label{comp1}
\end{figure}

\begin{figure}[!t]
\centering
\includegraphics[width=0.5\linewidth]{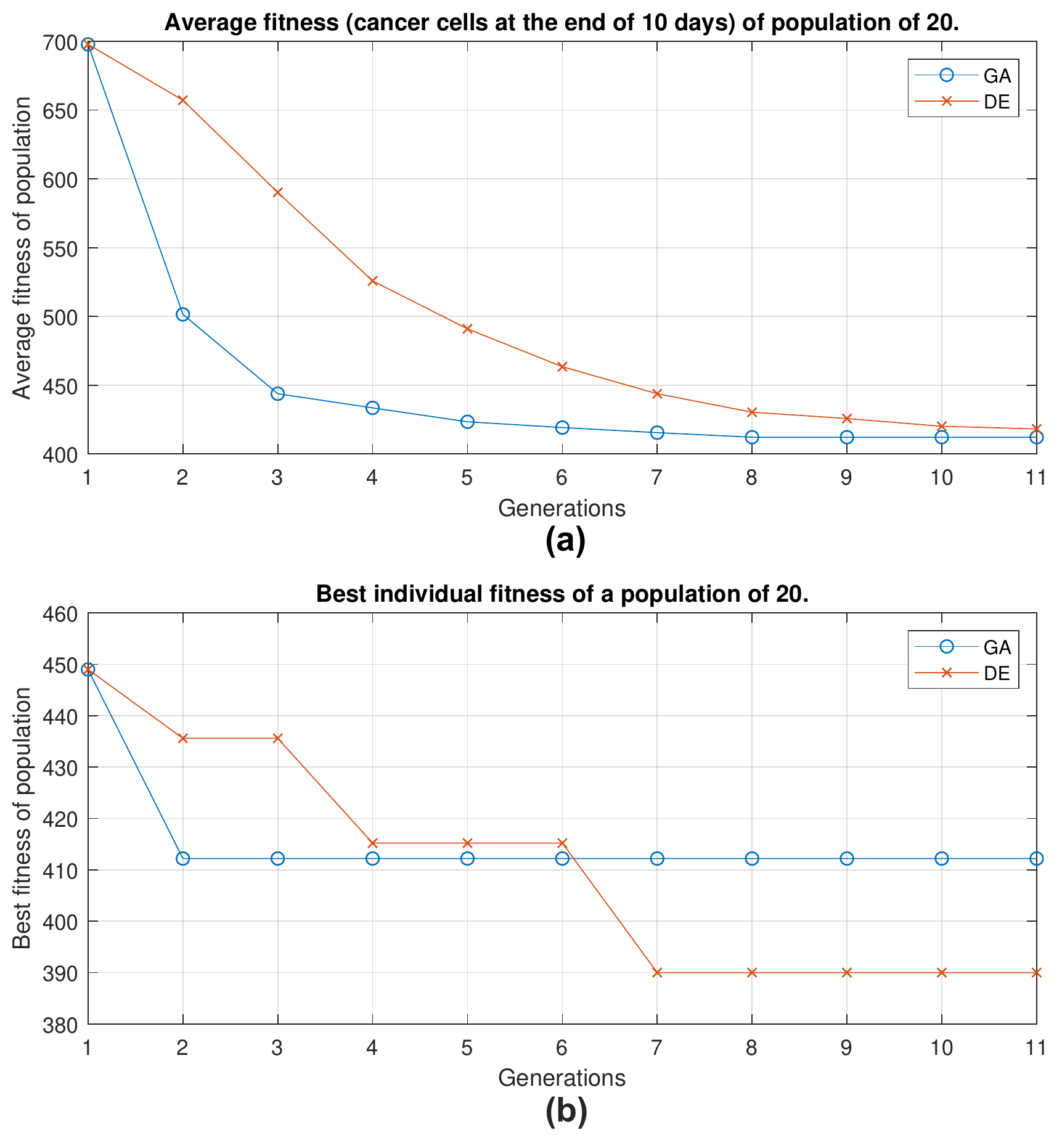}
\caption{Average (a) and best fitness (b) of the populations evolved with GA and DE in the second comparison run.}
\label{comp2}
\end{figure}

\begin{figure}[!t]
\centering
\includegraphics[width=0.5\linewidth]{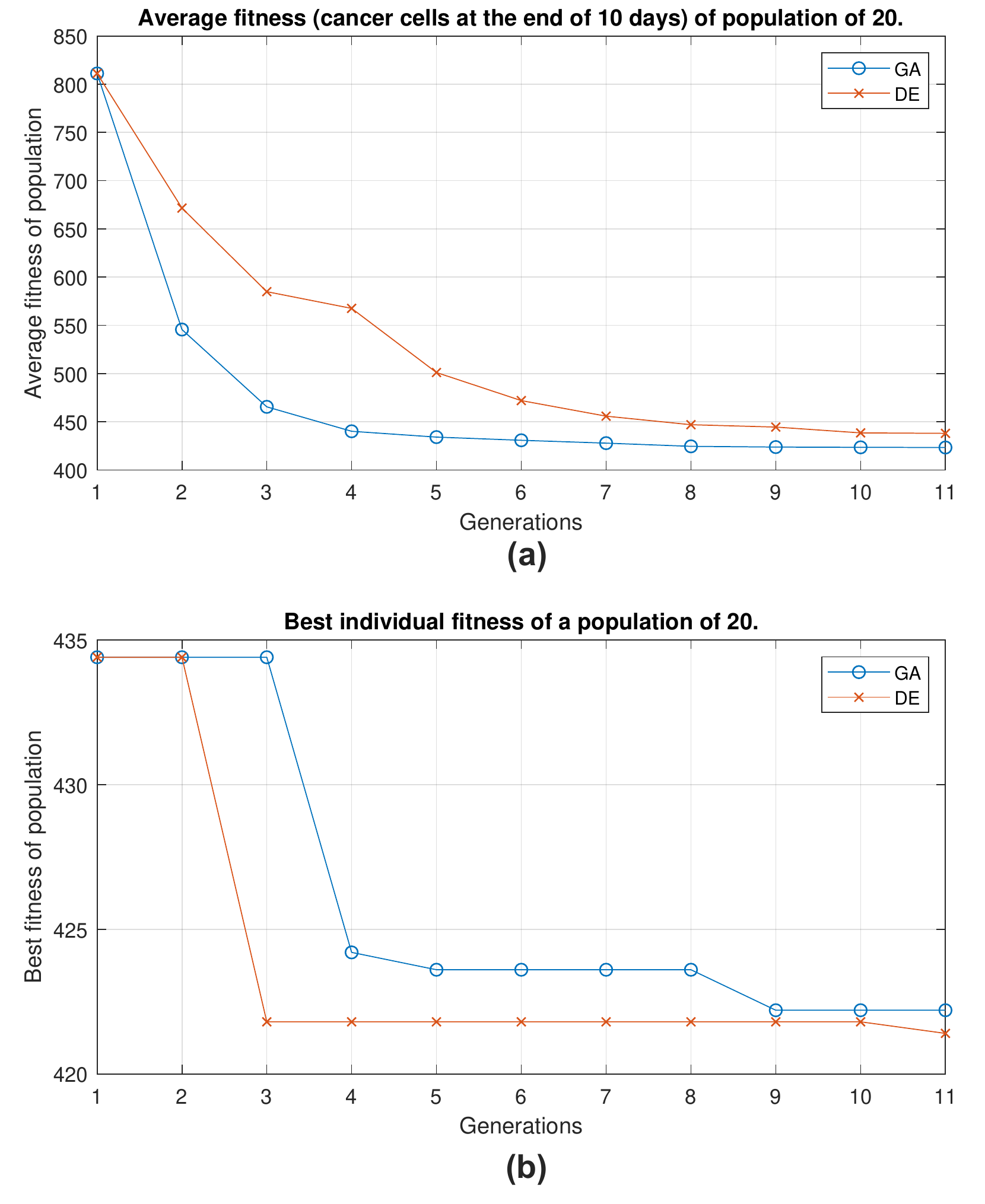}
\caption{Average (a) and best fitness (b) of the populations evolved with GA and DE in the third comparison run.}
\label{comp3}
\end{figure}

Three comparison tests were run, each using the same initial population for the GA and DE optimization. The results of the average fitness (remaining cancer cells after 10 days of simulation) of all individuals of the population through out the generations are depicted in Figs. \ref{comp1} - \ref{comp3}(a). Furthermore, the fitness of the best individual for both GA and DE throughout the generations was illustrated in Figs. \ref{comp1} - \ref{comp3}(b).

In Fig. \ref{comp1}(a) it is apparent that both approaches force the population to converge towards a lower fitness. However, in Fig. \ref{comp1}(b) the fact that DE outperforms the GA in finding better solutions from the first few generations is shown. Moreover, the GA seems to be stuck from the sixth generation in a local minimum.

In Fig. \ref{comp2}(a) the GA approach seems to converge the average fitness of the population of solutions towards a lower fitness faster than the DE. Also, in Fig. \ref{comp2}(b) the DE approach is outperformed by GA for the first six generations. On the contrary, the GA is again stuck in a local optimum, while the DE is continuously evolving towards better solutions and manages to find a better one at the sixth generation.

Finally, the results from the third comparison run are presented in Fig. \ref{comp3}. As in the previous runs, it can be claimed that DE outperforms the GA. Specifically, in Fig. \ref{comp3}(b) the DE approach reaches a fittest individual than GA at the very first generations. Furthermore, whereas the DE seems to be stuck in a local minimum from the third generation and the GA reaches a solution quite similar to this one, on the last generation the DE manages to escape its minimum and provide an ever better solution.

\begin{figure}[!t]
\centering
\includegraphics[width=0.5\linewidth]{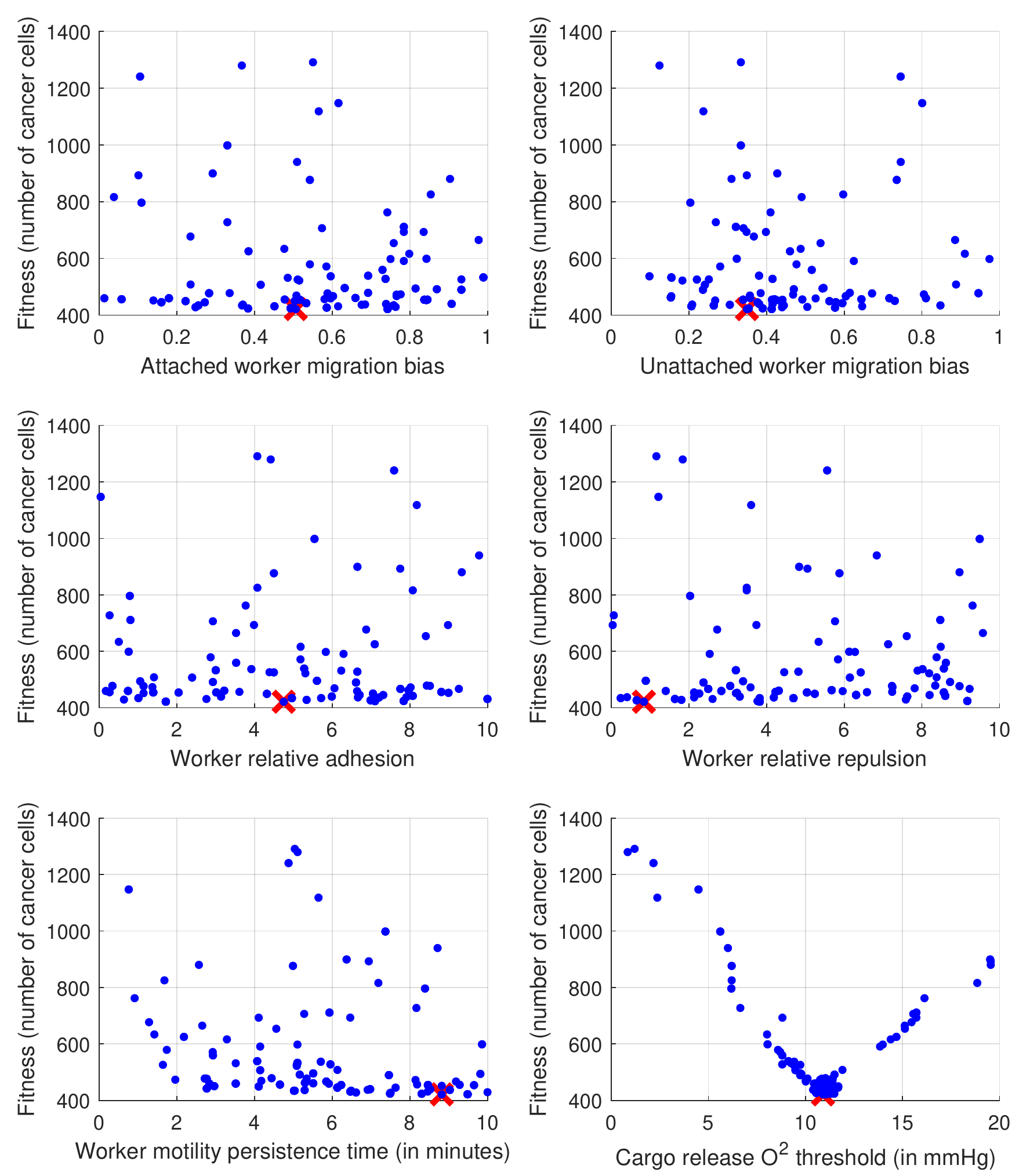}
\caption{Scatter plot of all individuals tested during the DE approach for the third run. The red ``X" mark denotes the best individual found.}
\label{par3}
\end{figure}

\begin{figure}[!t]
\centering
\includegraphics[width=0.5\linewidth]{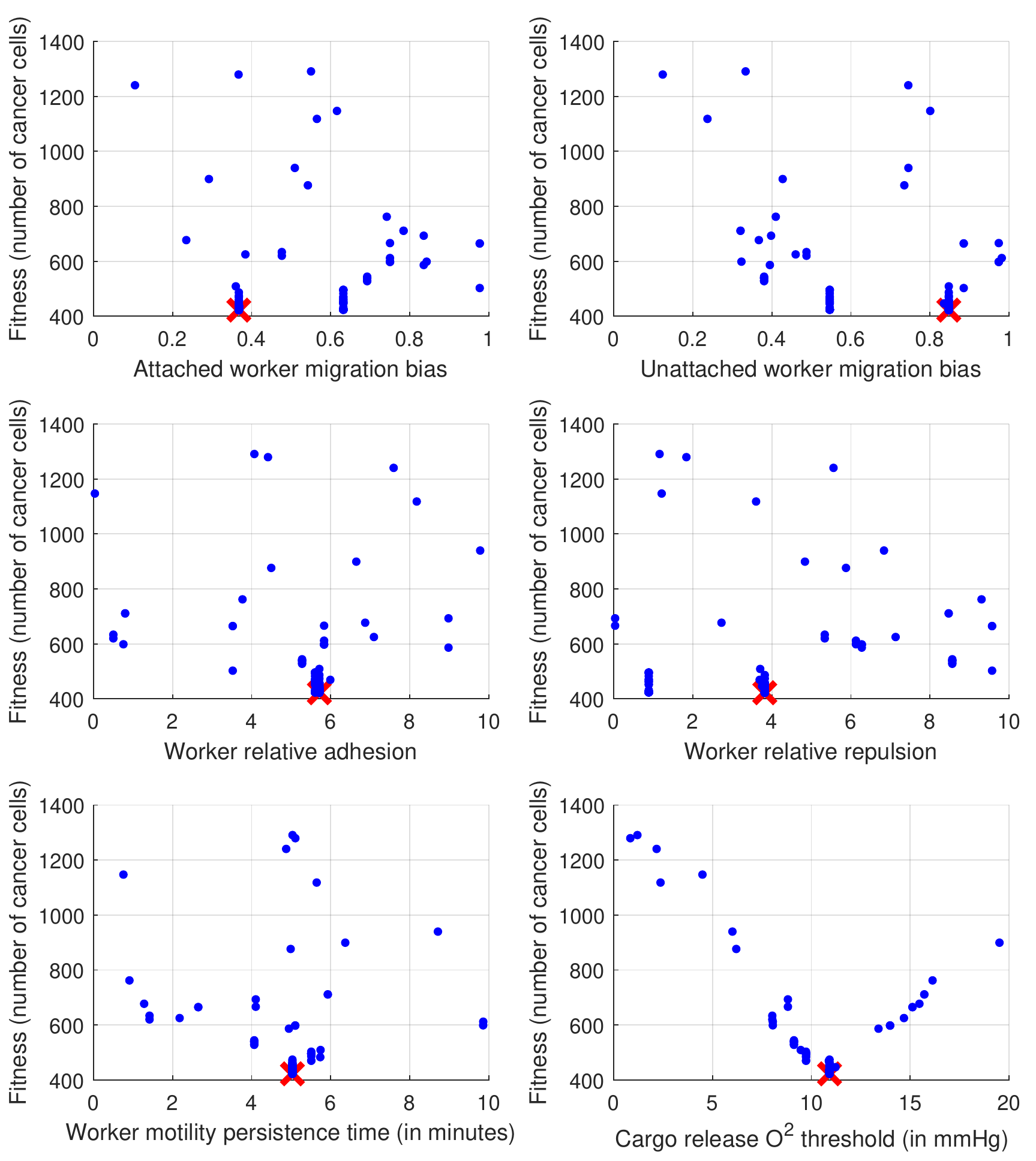}
\caption{Scatter plot of all individuals tested during the GA approach for the third run. The red ``X" mark denotes the best individual found.}
\label{par3all}
\end{figure}

The scatter plots of the parameters investigated during the evolution of DE in the third comparison run are outlined in Fig. \ref{par3}. %(for the first run in Fig. \ref{par1} and for the second run in Fig. \ref{par2} ... this can be added as Supplementary material to keep the paper small). 
It is clear that the individuals produced with the DE approach cover the search space better than the ones produced by GA (illustrated in Fig. \ref{par3all}). This is attributed to the fact that DE is designed to tackle models defined in real-values search spaces, whereas GA is not. As a result, the GA is heavily limited by the randomly produced initial population, a disadvantage that is alleviated by the DE methodology.

\begin{figure}[!t]
\centering
\includegraphics[width=0.5\linewidth]{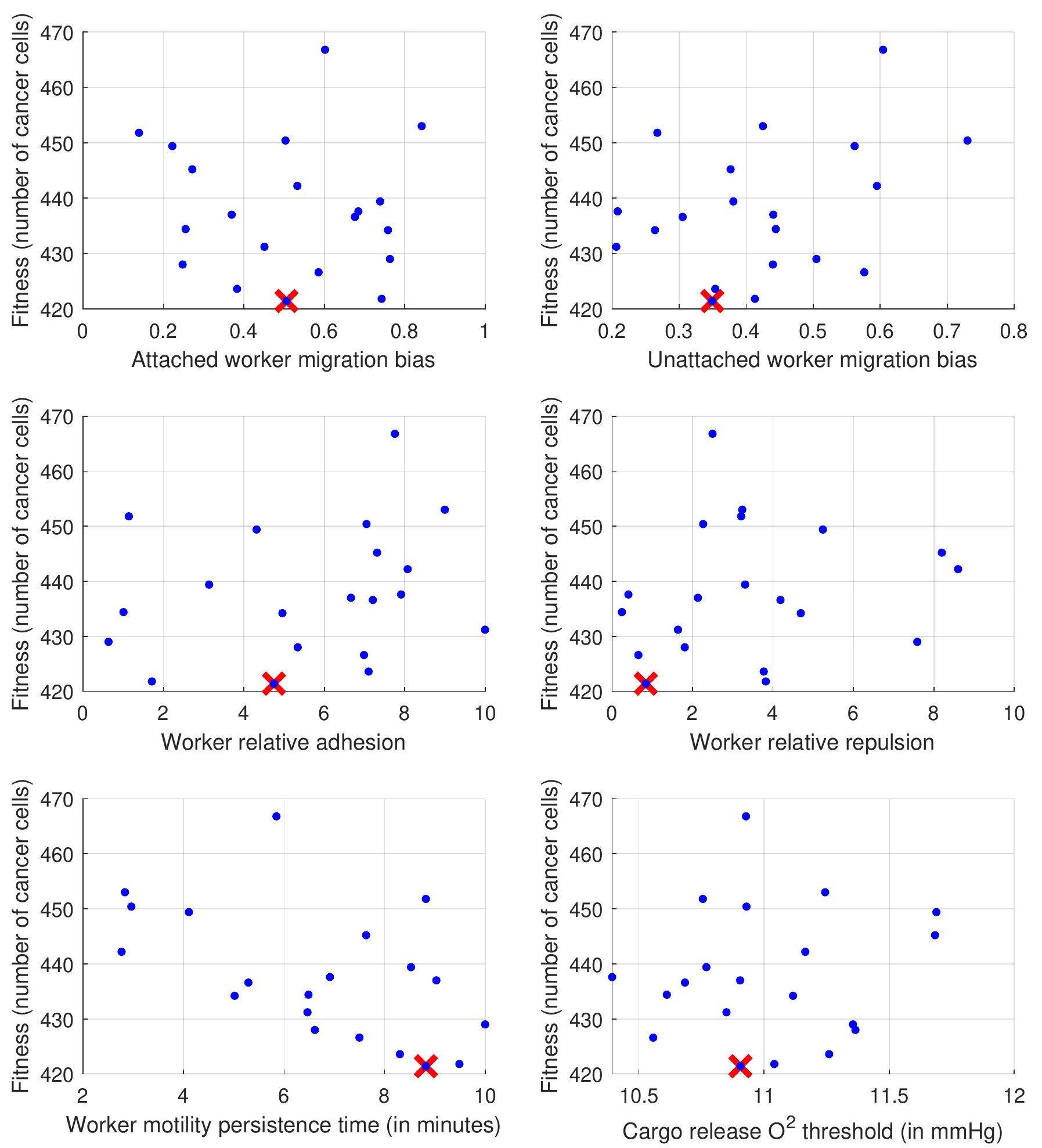}
\caption{Scatter plot of the individuals in the final population for the DE approach for the third run. The red ``X" mark denotes the best individual found.}
\label{par3finDE}
\end{figure}

\begin{figure}[!t]
\centering
\includegraphics[width=0.5\linewidth]{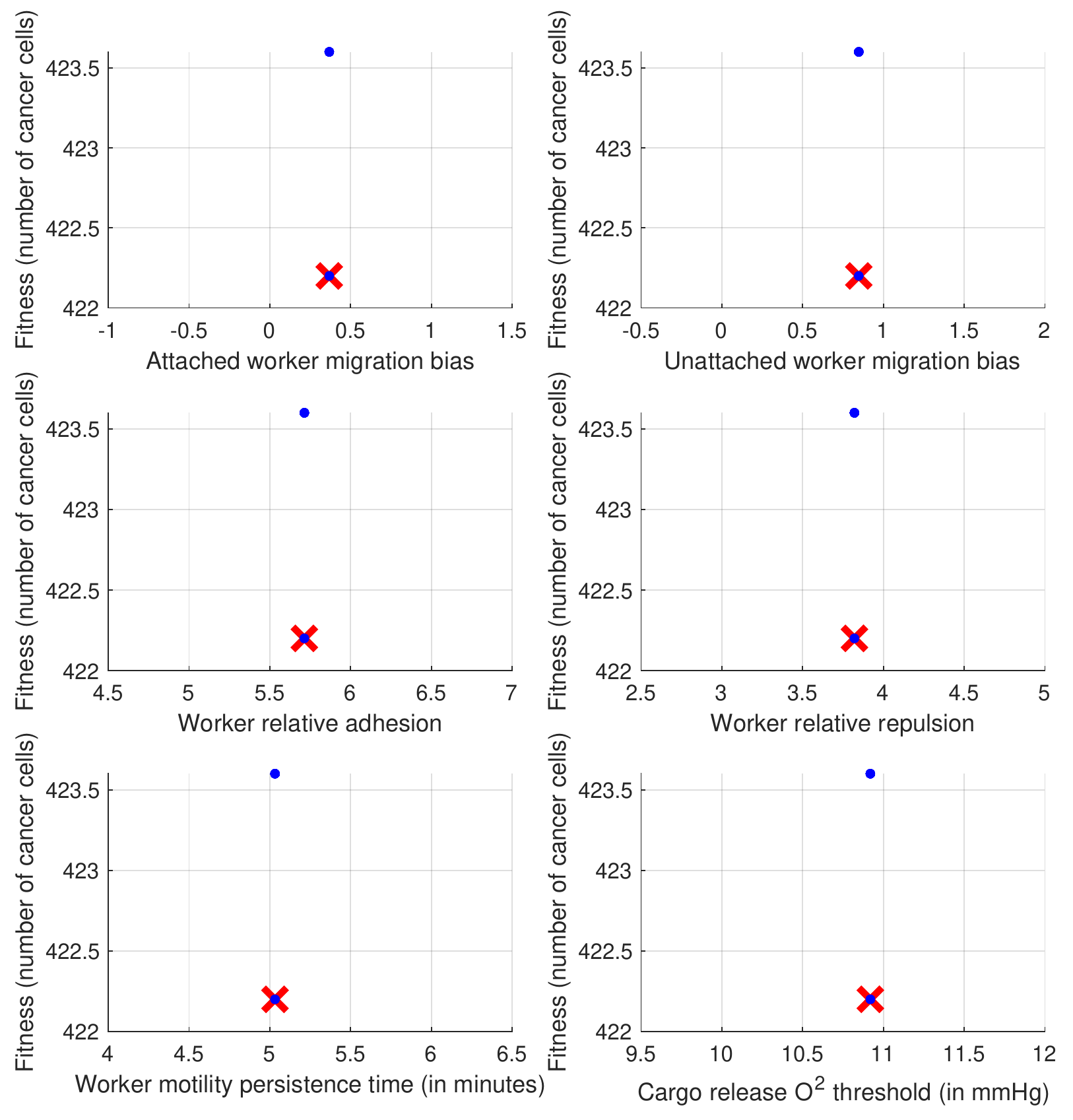}
\caption{Scatter plot of the individuals in the final population for the GA approach for the third run. The red ``X" mark denotes the best individual found.}
\label{par3fin}
\end{figure}

In addition, to clearly portray the reason why GA is easily trapped in local optima, while DE manages to escape them, Figs. \ref{par3finDE} and \ref{par3fin} are given, that present the final population of the DE and GA approaches for the third run, respectively. The DE approach succeeds in maintaining a high diversity in the final population as shown in Fig. \ref{par3finDE}, where no duplicate individuals can be spotted. On the other hand, in Fig. \ref{par3fin} the final population of the GA approach is apparently comprised by multiple copies of just two individuals, which are also very close to each other. Consequently, the GA can not escape from these two individuals unless a dramatic mutation happens (not possible as the mutation step size was set here to $s=[-5,5]\%$).

\section{Conclusion}
\label{S:5}

The optimization of the design of targeted DDS on a cancer tumor, simulated by PhysiCell, was studied by utilizing ``DE/rand/1" approach. The DE approach was compared with a standard steady state GA with the same computation budget, namely 1000 evaluations. The results derived from the comparison runs of both approaches equipped with the same initial population unveiled that DE is a more robust algorithm to reach a better solution within the same amount of generations. On top of that, DE enables the further exploration of the search space as it maintains a high diversity of the individuals in the final population.

As an aspect of future work, different variations of DE can be investigated with PhysiCell and under alternative ultimate goals, for instance the ability of niching is well documented in variants of DE \cite{epitropakis2013dynamic}. Finally, the conclusions driven from this study will be applied on ongoing research towards a more wide applicability platform that will design, develop and evaluate DDSs aiming cancer tumours.

\section{Acknowledgement}

This work was supported by the European Research Council under the European Union's  under grant agreement No. 800983.

\bibliographystyle{unsrt}

\begin{thebibliography}{10}

\bibitem{storn1997differential}
Rainer Storn and Kenneth Price.
\newblock Differential evolution--a simple and efficient heuristic for global
  optimization over continuous spaces.
\newblock {\em Journal of global optimization}, 11(4):341--359, 1997.

\bibitem{thomsen2004multimodal}
Rene Thomsen.
\newblock Multimodal optimization using crowding-based differential evolution.
\newblock In {\em Proceedings of the 2004 Congress on Evolutionary Computation
  (IEEE Cat. No. 04TH8753)}, volume~2, pages 1382--1389. IEEE, 2004.

\bibitem{epitropakis2011enhancing}
Michael~G Epitropakis, Dimitris~K Tasoulis, Nicos~G Pavlidis, Vassilis~P
  Plagianakos, and Michael~N Vrahatis.
\newblock Enhancing differential evolution utilizing proximity-based mutation
  operators.
\newblock {\em IEEE Transactions on Evolutionary Computation}, 15(1):99--119,
  2011.

\bibitem{qin2008differential}
A~Kai Qin, Vicky~Ling Huang, and Ponnuthurai~N Suganthan.
\newblock Differential evolution algorithm with strategy adaptation for global
  numerical optimization.
\newblock {\em IEEE transactions on Evolutionary Computation}, 13(2):398--417,
  2008.

\bibitem{das2010differential}
Swagatam Das and Ponnuthurai~Nagaratnam Suganthan.
\newblock Differential evolution: A survey of the state-of-the-art.
\newblock {\em IEEE transactions on evolutionary computation}, 15(1):4--31,
  2010.

\bibitem{DAS20161}
Swagatam Das, Sankha~Subhra Mullick, and P.N. Suganthan.
\newblock Recent advances in differential evolution - {A}n updated survey.
\newblock {\em Swarm and Evolutionary Computation}, 27:1 -- 30, 2016.

\bibitem{wang2013gaussian}
Hui Wang, Shahryar Rahnamayan, Hui Sun, and Mahamed~GH Omran.
\newblock Gaussian bare-bones differential evolution.
\newblock {\em IEEE Transactions on Cybernetics}, 43(2):634--647, 2013.

\bibitem{price1997differential}
Kenneth~V Price.
\newblock Differential evolution vs. the functions of the 2/sup nd/iceo.
\newblock In {\em Proceedings of 1997 IEEE International Conference on
  Evolutionary Computation (ICEC'97)}, pages 153--157. IEEE, 1997.

\bibitem{ghaffarizadeh2018physicell}
Ahmadreza Ghaffarizadeh, Randy Heiland, Samuel~H Friedman, Shannon~M
  Mumenthaler, and Paul Macklin.
\newblock Physicell: an open source physics-based cell simulator for 3-d
  multicellular systems.
\newblock {\em PLoS computational biology}, 14(2):e1005991, 2018.

\bibitem{ghaffarizadeh2015biofvm}
Ahmadreza Ghaffarizadeh, Samuel~H Friedman, and Paul Macklin.
\newblock Biofvm: an efficient, parallelized diffusive transport solver for 3-d
  biological simulations.
\newblock {\em Bioinformatics}, 32(8):1256--1258, 2015.

\bibitem{PREEN20191}
Richard~J. Preen, Larry Bull, and Andrew Adamatzky.
\newblock Towards an evolvable cancer treatment simulator.
\newblock {\em Biosystems}, 182:1 -- 7, 2019.

\bibitem{tsompanas2019}
Michail-Antisthenis Tsompanas, Larry Bull, Andrew Adamatzky, and Igor Balaz.
\newblock Haploid-diploid evolution: Nature's memetic algorithm, 2019.

\bibitem{ozik2019learning}
Jonathan Ozik, Nicholson Collier, Randy Heiland, Gary An, and Paul Macklin.
\newblock Learning-accelerated discovery of immune-tumour interactions.
\newblock {\em Molecular Systems Design \& Engineering}, 2019.

\bibitem{bull1999baldwin}
Larry Bull.
\newblock On the baldwin effect.
\newblock {\em Artificial life}, 5(3):241--246, 1999.

\bibitem{bull2017evolution}
Larry Bull.
\newblock The evolution of sex through the baldwin effect.
\newblock {\em Artificial life}, 23(4):481--492, 2017.

\bibitem{epitropakis2013dynamic}
Michael~G Epitropakis, Xiaodong Li, and Edmund~K Burke.
\newblock A dynamic archive niching differential evolution algorithm for
  multimodal optimization.
\newblock In {\em 2013 IEEE Congress on Evolutionary Computation}, pages
  79--86. IEEE, 2013.

\end{thebibliography}

\end{document}